\documentclass[letterpaper]{article} 
\usepackage{aaai2026}  
\usepackage{times}  
\usepackage{helvet}  
\usepackage{courier}  
\usepackage[hyphens]{url}  
\usepackage{graphicx} 
\usepackage{natbib}  
\usepackage{caption} 
\usepackage{algorithm}
\usepackage{algorithmic}

\usepackage{booktabs} 
\usepackage{amsmath}
\usepackage{amsfonts}
\usepackage{multirow}
\usepackage{array}
\usepackage{subcaption}

\usepackage{newfloat}
\usepackage{listings}
\DeclareCaptionStyle{ruled}{labelfont=normalfont,labelsep=colon,strut=off} 
\floatstyle{ruled}
\newfloat{listing}{tb}{lst}{}
\floatname{listing}{Listing}
\title{Resilience Inference for Supply Chains with Hypergraph Neural Network}
\author{
    Zetian Shen\textsuperscript{\rm 1}\equalcontrib,
    Hongjun Wang\textsuperscript{\rm 2,4}\equalcontrib,
    Jiyuan Chen\textsuperscript{\rm 3,4},
    Xuan Song\textsuperscript{\rm 1}\thanks{Corresponding Author}
}
\affiliations{
    \textsuperscript{\rm 1}School of Artificial Intelligence, Jilin University\\
    \textsuperscript{\rm 2}The University of Tokyo \\
    \textsuperscript{\rm 3}Hong Kong Polytechnic University \\
    \textsuperscript{\rm 4}Southern University of Science and Technology\\
}

\begin{document}

\maketitle

\begin{abstract}
Supply chains are integral to global economic stability, yet disruptions can swiftly propagate through interconnected networks, resulting in substantial economic impacts. Accurate and timely inference of supply chain resilience—the capability to maintain core functions during disruptions—is crucial for proactive risk mitigation and robust network design. However, existing approaches lack effective mechanisms to infer supply chain resilience without explicit system dynamics and struggle to represent the higher-order, multi-entity dependencies inherent in supply chain networks. These limitations motivate the definition of a novel problem and the development of targeted modeling solutions. To address these challenges, we formalize a novel problem: Supply Chain Resilience Inference (SCRI), defined as predicting supply chain resilience using hypergraph topology and observed inventory trajectories without explicit dynamic equations. To solve this problem, we propose the Supply Chain Resilience Inference Hypergraph Network (SC-RIHN), a novel hypergraph-based model leveraging set-based encoding and hypergraph message passing to capture multi-party firm-product interactions. Comprehensive experiments demonstrate that SC-RIHN significantly outperforms traditional MLP, representative graph neural network variants, and ResInf baselines across synthetic benchmarks, underscoring its potential for practical, early-warning risk assessment in complex supply chain systems.
\end{abstract}

\begin{links}
    \link{Extended version}{https://aaai.org/example}
\end{links}

        \section{Introduction}

\begin{figure}[t]
        \centering
        \includegraphics[width=0.98\columnwidth]{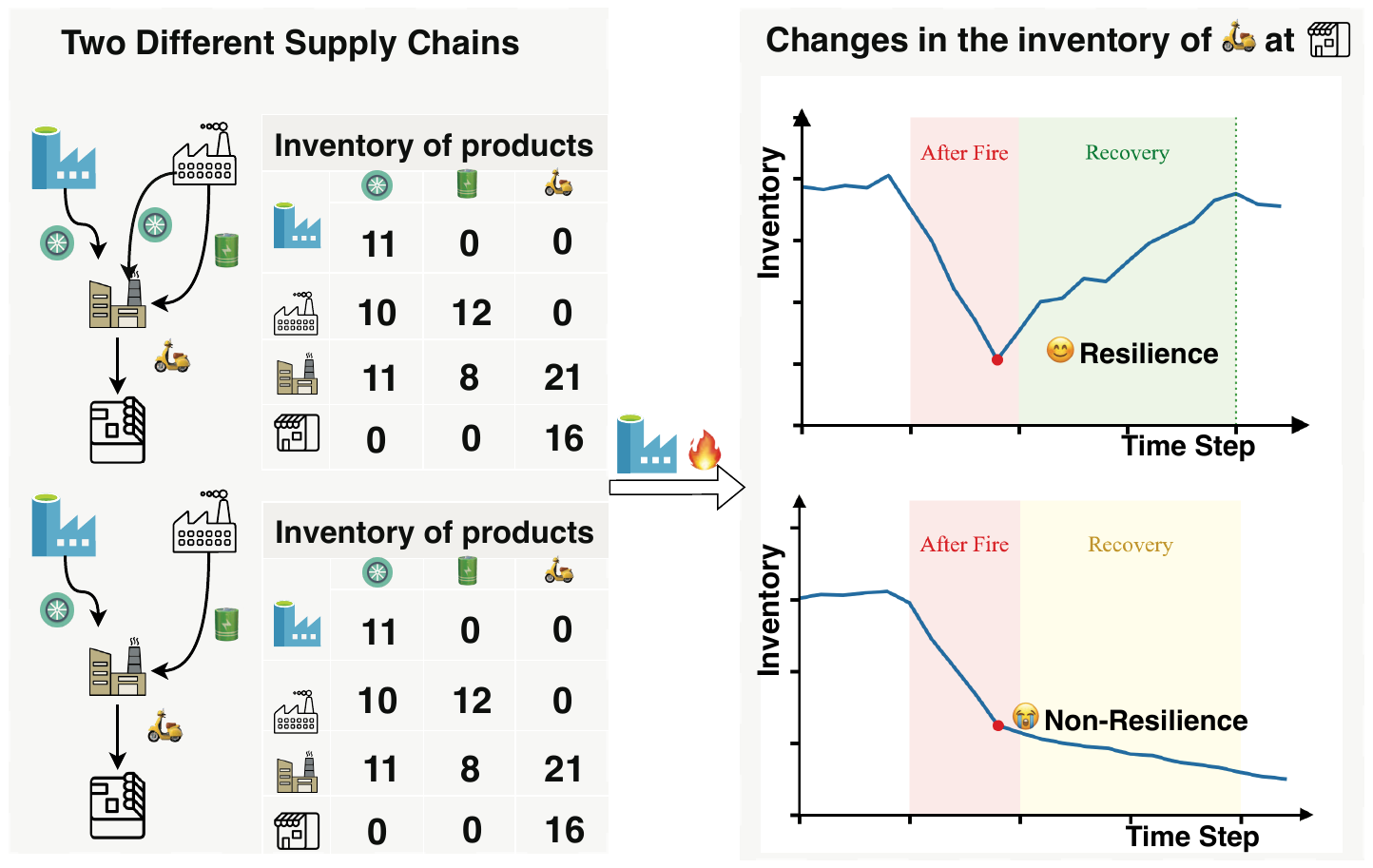}
        \caption{Illustration of supply chain resilience. Two supply chains with different topologies (left) experience the same disruption (a factory fire). The inventory trajectory of a downstream store (right) shows recovery in the upper network (resilient) and persistent failure in the lower network (non-resilient).}
        \label{fig:SCR-intro}
    \end{figure}

    Supply chains are critical to the global economy, yet localized disruptions—such as production accidents, logistics delays, or extreme weather—can propagate through interconnected networks, resulting in massive economic losses and even posing serious risks to national security~\cite{baumgartner2020reimagining}. However, supply chains differ significantly in their resilience to such shocks, where resilience is defined as the ability of a system to adapt and maintain core functions under disruption~\cite{cohen2000resilience, paper1, liu2022network}. Figure~\ref{fig:SCR-intro} illustrates two similarly sized supply chains subjected to the same factory fire, where differences in network topology and product dependencies result in rapid recovery for one and prolonged stagnation for the other. These observations raise a key question: can we achieve early inference of supply chain resilience to support proactive risk mitigation and robust network design?

    Traditional supply chain studies relied on deterministic network optimisation~\cite{vidal1997strategic}, system‑dynamics simulation~\cite{sterman1989misperceptions}, and agent‑based modelling~\cite{wu2012supply}, while resilience work examined buffering and redundancy through stochastic or queuing frameworks~\cite{sheffi2005supply,ivanov2018structural}. However, as global supply chains grow in complexity and scale, these mechanistic models struggle to fit even highly aggregated measures like country-level production~\cite{inoue2019firm}, motivating the adoption of data-driven approaches such as machine learning for risk assessment and disruption prediction~\cite{baryannis2019predicting,brintrup2020supply}. Moreover, graph neural networks (GNNs) have shown strong potential for representing firm interactions and capturing the structural properties of supply chain networks, with applications including hidden‑link recovery, supplier risk ranking, and production‑function estimation~\cite{aziz2021data,kosasih2022machine,wasi2024supplygraph,chang2025learning}. Despite these advances, data‑driven approaches aimed specifically at supply chain resilience inference remain limited.

    Meanwhile, GBB~\cite{paper1} extended resilience analysis from low‑dimensional models to interacting complex networks, providing an important theoretical foundation for resilience prediction. More recently, GNN-based ResInf~\cite{liu2022network} inferred resilience directly from graph structure and observed state trajectories, bypassing explicit modeling of system dynamics while demonstrating strong performance in mutualistic, gene regulatory, and neuronal networks. Nevertheless, supply chains differ fundamentally: they exhibit higher-order \textit{firm–product–firm} structures where firms are linked through shared inputs and outputs~\cite{carvalho2019production, chang2025learning}. Such higher-order dependencies are naturally modeled as hypergraphs, where a hyperedge captures upstream–downstream relations via shared products. Standard GNNs, which operate on binary relations between nodes, are insufficient to model these multi-party dependencies. Thus, it is necessary to develop models that can effectively capture higher-order dependencies inherent in hypergraph-structured supply chains.

    Consequently, two open challenges remain for supply chain resilience inference. At the algorithmic level, existing approaches lack a learnable mechanism to infer supply chain resilience directly from historical inventory trajectories without explicit knowledge of the underlying system dynamics. At the structural level, GNN-based models struggle to represent the higher-order dependencies and multi-entity interactions that are inherent in supply chain networks, limiting their applicability for resilience inference.

    Motivated by these challenges, we identify a challenging unexplored problem: \emph{Supply Chain Resilience Inference (SCRI)}, which aims to predict the resilience of a supply chain based on its hypergraph topology and observed state trajectories. SCRI focuses on predicting supply chain resilience to enable early-warning capabilities, but is challenged by the absence of explicit system dynamics and the complexity of the underlying hypergraph structure. To address SCRI, we propose the Supply‑Chain Resilience Inference Hypergraph Network (SC‑RIHN), a model that leverages hypergraph message passing to learn higher-order interaction patterns from the supply chain structure and historical state trajectories, and integrates these representations into a resilience embedding for inference. Experiments on both synthetic shock scenarios and real-world datasets show that SC‑RIHN consistently outperforms MLP, representative GNN variants, and the ResInf baseline across standard metrics. In summary, our main contributions are:
    \begin{enumerate}
        \item We formalise \emph{Supply Chain Resilience Inference}, a new problem of predicting supply chain resilience from hypergraph topology and historical state trajectories.
        \item We propose SC‑RIHN, a hypergraph network that captures higher-order interactions through hypergraph message passing.
        \item  We provide a process for constructing a synthetic benchmark and generating resilience labels to support reproducible evaluation.
        \item The experimental results indicate that SC-RIHN outperforms representative GNN baselines by more effectively capturing higher-order interactions, underscoring its potential for early-warning risk assessment.
    \end{enumerate}

    \section{Related Work}

    Supply chains are often modeled as complex networks, where firms serve as nodes and supply or trade relationships form the edges. This network-based representation has facilitated quantitative analyses of systemic risk propagation~\cite{fujiwara2010large,acemoglu2012network,zhao2019modelling,carvalho2021supply}. In this context, resilience analysis typically employs system dynamics simulations or stochastic models to evaluate recovery following disruptions~\cite{hallegatte2008adaptive,guan2020global,sheffi2005supply,ivanov2018structural}. However, these methods frequently depend on extensive domain-specific assumptions and exhibit poor scalability in realistic, high-dimensional scenarios~\cite{inoue2019firm}.

    Graph neural networks (GNNs) have recently been employed in supply chain systems, leveraging their inherent graph structure for tasks such as prediction and optimization~\cite{aziz2021data,kosasih2022machine, ahn2024gnn}. The \textsc{SupplyGraph} benchmark~\cite{wasi2024supplygraph} addresses the lack of real-world datasets by providing temporal supply chain data from a major FMCG firm, enabling GNN-based modeling of sales, production, and factory issues. To address data privacy and distribution challenges, a federated GNN framework~\cite{qu2023flee} enables decentralized analysis of geospatial resilience in multicommodity food flow networks. Additionally, production function inference in firm–product networks has been enhanced using GNNs equipped with temporal encoding and inventory-aware modules, leading to notable improvements in supply forecasting~\cite{chang2025learning}. While these approaches advance task-specific prediction, they fall short of inferring system-level resilience or explicitly modeling higher-order supply chain dependencies.

    To overcome the limitations of GNNs in modeling higher-order relationships, hypergraph neural networks (HGNNs) extend GNNs by introducing hyperedges that connect arbitrary sets of nodes, enabling more expressive group-level representations~\cite{feng2019hypergraph,gao2022hgnn+,li2025deep}. For example, a neuromodulated small-world HGNN enhances trajectory prediction by capturing both local and long-range vehicle interactions, showcasing the capability of HGNNs in modeling complex, higher-order systems~\cite{wang2025nest}. Despite their potential, the application of hypergraph-based learning in supply chain systems remains limited. Some initial efforts have explored firm–product–firm hypergraphs to encode higher-order relationships~\cite{chang2025learning}, but none have focused on resilience inference.

Classical complexity–stability theory~\cite{may1972will, paper1} and its graph-neural extension ResInf~\cite{liu2022network} estimate resilience directly from standard graph structures. However, supply chains inherently form higher-order \textit{firm–product–firm} hypergraphs~\cite{carvalho2019production}, underscoring the need for hypergraph-based resilience inference.

    \section{Preliminaries}\label{sec:preliminaries}

This section introduces key concepts and notations for supply chain modeling, resilience, and the Supply Chain Resilience Inference (SCRI) problem.

    \subsection{Supply Chain and Temporal States}\label{subsec:supply-chain-and-temporal-states}
    We represent the supply chain as a tripartite hypergraph \(\mathcal{H} = (\mathcal{C}, \mathcal{P}, \mathcal{E})\), where \(\mathcal{C}\) denotes the set of firm nodes, \(\mathcal{P}\) is the set of product nodes, and \(\mathcal{E} \subseteq \mathcal{C} \times \mathcal{P} \times \mathcal{C}\) represents the set of hyperedges. Each hyperedge \((c_{\text{up}}, p, c_{\text{down}}) \) indicates that the downstream firm \(c_{\text{down}}\) procures product \(p\) from the upstream firm \(c_{\text{up}}\). Additionally, each firm node $c$ is associated with a time-dependent state vector $\smash{\mathbf{x}_c^{(t)}} \in \mathbb{R}^d$, capturing its operational status at time $t$. The dimensions of this vector may include key performance indicators such as inventory levels, production rates, and order volumes. In this work, we use inventory vectors to define node states: $\mathbf{x}_c^{(t)} = \bigl(I_{c,p}^{(t)}\bigr)_{p \in \mathcal{P}} \in \mathbb{R}^{|\mathcal{P}|}$, where $I_{c,p}^{(t)}$ denotes the inventory level of product $p$ held by firm $c$ at time $t$. Stacking these vectors across all firms forms the supply chain state matrix $\mathbf{X}_t \in \mathbb{R}^{|\mathcal{C}| \times |\mathcal{P}|}$.

    \subsection{Resilience}\label{subsec:resilience}

    \paragraph{Complex Network Resilience.}

    Consider a complex networked system $G = (V, A)$, where $A$ is the  adjacency matrix, $V$ is the set of nodes and each node $i \in V$ has a state $x_i$. The system evolves according to the dynamics:
    \begin{equation}
        \frac{dx_i}{dt} = F(x_i) + \sum_{j=1}^{N} A_{ij}\,G(x_i, x_j)
        \label{eq:dynamics}
    \end{equation}
    where $F(x_i)$ captures the intrinsic behavior of node $i$, and $G(x_i, x_j)$ encodes the interaction between nodes $i$ and $j$. As defined in GBB~\cite{paper1}, the system is \emph{resilient} if it has a unique and stable equilibrium $x^* \neq 0$ under the dynamics in Equation~\eqref{eq:dynamics}. In such systems, any bounded perturbation to the states $x_i$ will decay over time, leading the system to converge to $x^*$. Conversely, a \emph{non-resilient} system either fails to return to a desirable state or diverges entirely.

    \paragraph{Supply chain Resilience.}
    Similar to the ResInf framework~\cite{ResInf}, we define supply chain resilience in terms of the convergence of firm-level states. Specifically, the primary concern lies in the operational states of firms, such as inventory levels and production capacities, rather than product-level dynamics. Formally, let $\mathbf{X}_t = \{\mathbf{x}_c^{(t)}\}_{c \in \mathcal{C}}$ denote the collection of firm states at time $t$. The supply chain is considered \emph{resilient} if $\mathbf{X}^{(t)}$ converges to a unique, non-zero equilibrium $\mathbf{X}^*$ from any feasible initial condition. Otherwise, the system is \emph{non-resilient} if its dynamics remain unstable or fail to converge to a stable equilibrium.

    \subsection{Problem Definition}\label{subsec:problem-definition}

    We define Supply Chain Resilience Inference (SCRI) as the problem of inferring whether a supply chain is resilient based on its network structure and historical inventory dynamics. Formally, given a supply chain hypergraph \(\mathcal{H}\) and a historical observation window \(\mathcal{W}_{T} = [\mathbf{X}_1, \mathbf{X}_2, \dots, \mathbf{X}_{T-1}] \in \mathbb{R}^{T \times |\mathcal{C}| \times D_\text{in}}\), where \(\mathbf{X}_t\) is the system state matrix at time \(t\), $T$ denotes the total number of time steps, and \(D_\text{in}\) represents the dimension of node states. In this work, node states are represented by inventory vectors, so that $D_\text{in} = |\mathcal{P}|$ denotes the number of products, potentially varying across supply chains.
    The task of the SCRI is to learn a parameterized mapping \(f_\theta : (\mathcal{W}_T, \mathcal{H}) \to \{0,1\}\) that predicts whether the system is resilient. This formulation infers resilience based on observed trajectories and the supply chain structure, without requiring explicit modeling of the underlying dynamics.

\begin{figure*}[t]
        \centering
        \includegraphics[width=0.98\textwidth]{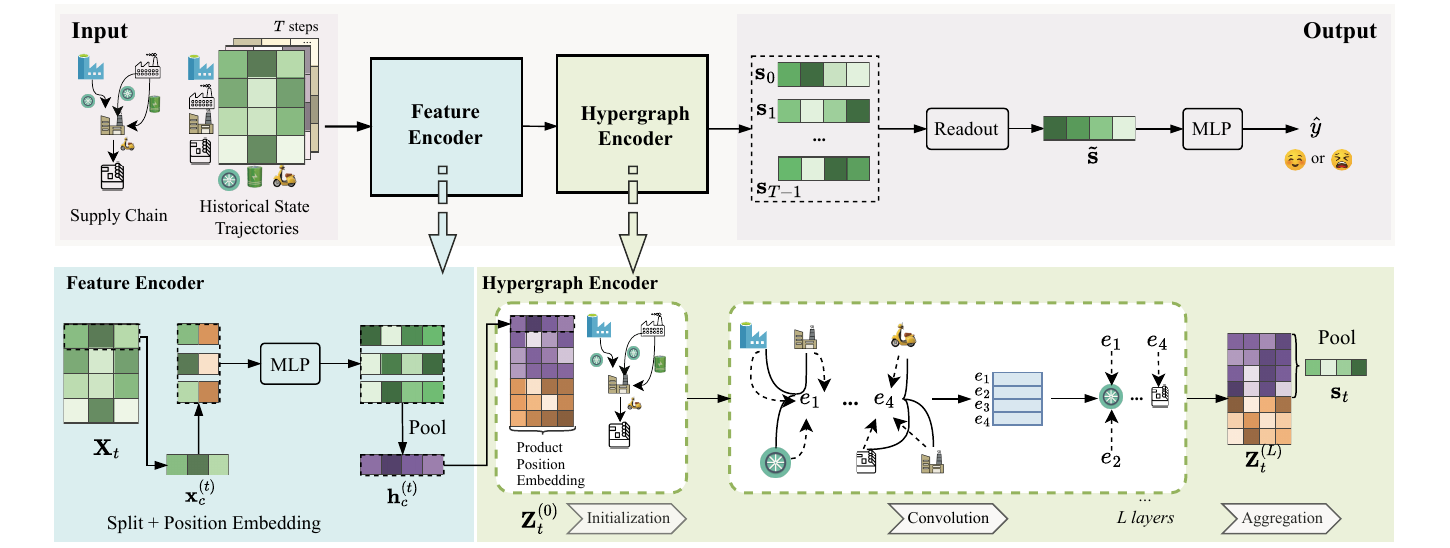}
        \caption{Overview of the proposed Supply Chain Resilience Inference Hypergraph Network (SC-RIHN). At each time step, the Feature Encoder decomposes firm state vectors into individual feature dimensions and combines them with learnable positional embeddings. A multi-layer perceptron (MLP) followed by a pooling operation then projects these heterogeneous inputs from various supply chains into a unified latent space. Subsequently, the Hypergraph Encoder initializes product node features using positional embeddings and integrates them with firm node embeddings. After multiple hypergraph convolution layers capturing higher-order dependencies, the refined firm node embeddings are pooled into a graph-level representation. Finally, the Global Readout layer aggregates these representations across all time steps into a single system-level resilience embedding, which an MLP uses to infer resilience.}
        \label{fig:method-overview}
    \end{figure*}

    \section{The Proposed Method}\label{sec:the-proposed-method}

We propose a novel framework, termed Supply Chain Resilience Inference Hypergraph Network (SC-RIHN), to infer the resilience of supply chains by capturing higher-order interactions between firms and products. The overall architecture of SC-RIHN is illustrated in Figure~\ref{fig:method-overview}. Given a supply chain hypergraph and corresponding historical state trajectories, a feature encoder first transforms each firm's variable-length state vector into a fixed-dimensional embedding. This transformation ensures consistent representation across heterogeneous supply chains while preserving index-specific information essential for downstream inference.
The resulting embeddings are then processed by a hypergraph encoder, which propagates information through the hypergraph to capture multi-party dependencies. At each time step, structural representations are obtained by applying a pooling function over the updated embeddings of firm nodes. To obtain global system‑level information, a global readout layer aggregates the sequence of structural embeddings into a unified resilience representation, which is finally used  for resilience inference via a multi-layer perceptron (MLP). Moreover, the model is trained end-to-end using a binary cross-entropy loss to optimize resilience inference.

    \subsection{Feature Encoder}\label{subsec:feature-encoder}
    Firms are typically associated with feature sets whose dimensionality can vary across different supply chain networks. For instance, differences in the number of products handled by each supply chain result in inventory vectors of varying lengths. To address this variability, we employ a set-based feature encoder inspired by DeepSets~\cite{zaheer2017deep}, extending it with learnable positional embeddings to distinguish feature indices during aggregation. This design enables the model to process inputs of arbitrary dimensionality while preserving information about the identity and positional context of each feature.

    Specifically, at each time step \(t\), the feature encoder transforms the input feature vector $\mathbf{x}_c^{(t)}$ of firm \(c \in \mathcal{C}\) into a fixed-dimensional representation \(\mathbf{h}_c^{(t)} \in \mathbb{R}^{D_\text{hidden}}\), ensuring that firm nodes across different supply chains are represented within a consistent feature space. To achieve this, the input vector is initially split into individual feature dimensions, where each feature is combined with a learnable positional embedding that encodes its index. These representations are then passed through a shared multilayer perceptron (MLP), and the transformed vectors are aggregated to obtain the final firm-level embedding. Formally, the feature encoder is defined as:
    \begin{equation}
        \mathbf{h}_c^{(t)} = \mathrm{Pool}\left(\left\{\phi\left(x_{c,d}^{(t)}, \mathrm{pos}(d)\right) \mid d = 1, \dots, D_\text{in}\right\}\right)
        \label{eq:feature-encoder}
    \end{equation}
    where $x_{c,d}^{(t)}$ is the $d$-th feature of firm $c$ at time $t$,
$\mathrm{pos}(d)$ is a learnable positional embedding for feature index $d$, \(\phi(\cdot)\) is an MLP applied to each individual feature, and \(\mathrm{Pool}(\cdot)\) performs aggregation over the feature dimension (e.g., via summation or averaging) to produce a firm-level embedding.

    Although feature vectors are aggregated as sets for aggregation purposes, we introduce learnable positional embeddings \(\mathrm{pos}(d)\) to distinguish feature indices \(d\). This enables the model to capture structural patterns encoded in the feature ordering, which may reflect domain-specific semantics, such as the prioritization of certain attributes in supply chain operations.

    \subsection{Hypergraph Encoder}\label{subsec:hypergraph-encoder}

    To capture higher-order dependencies, we adopt hypergraph convolution operations on the supply chain hypergraph $\mathcal{H} = (\mathcal{C}, \mathcal{P}, \mathcal{E})$. The module first initializes node features for both firms and products, then propagates information through shared product nodes using hypergraph message passing, and finally aggregates firm embeddings to obtain a compact system-level representation. This design enables the model to capture complex interactions and multi-hop dependencies across different levels of the supply chain.

    \paragraph{Node Feature Initialization.}

    At each time step $t$, we construct an augmented node feature space that includes both firm and product nodes. For each firm node $c \in \mathcal{C}$, we assign a dynamic embedding $\mathbf{h}_c^{(t)}$ generated by the feature encoder. In contrast, product nodes $p \in \mathcal{P}$ are considered static, lacking intrinsic temporal dynamics. To encode their identity and enable structure-aware learning within the hypergraph, each product node is assigned a learnable positional embedding $\mathrm{pos}(p)$. Combining these representations, we define the initial feature matrix as $\mathbf{Z}_t^{(0)} = \mathrm{stack}(\{\mathbf{h}_c^{(t)}\}_{c \in \mathcal{C}}, \{\mathrm{pos}(p)\}_{p \in \mathcal{P}})$, where $\mathrm{stack}(\cdot)$ indicates vertical concatenation along the node dimension.

    \paragraph{Hypergraph Convolution.}
To model higher-order interactions between firms and products, we apply Hypergraph Neural Networks (HGNN)~\cite{feng2019hypergraph} for message passing over the supply chain hypergraph. Formally, the hypergraph convolution at the $l$-th layer is defined as:

\begin{equation}
    \mathbf{Z}_t^{(l+1)} = \sigma\left( \mathbf{D}_v^{-1/2} \mathbf{H} \mathbf{W}_e \mathbf{D}_e^{-1} \mathbf{H}^\top \mathbf{D}_v^{-1/2} \mathbf{Z}_t^{(l)} \mathbf{W}^{(l)} \right)
    \label{eq:hypergraph-convolution}
\end{equation}
where $\mathbf{Z}_t^{(l)}$ is the node feature matrix, $\mathbf{H}$ is the incidence matrix encoding node–hyperedge relationships, $\mathbf{W}^{(l)}$ is a learnable weight matrix, and $\sigma(\cdot)$ denotes a nonlinear activation function. The degree matrices $\mathbf{D}_v$ and $\mathbf{D}_e$ normalize the aggregation to ensure training stability. The hyperedge weight matrix $\mathbf{W}_e$ is initialized as the identity matrix, assigning equal weights to all hyperedges.

Conceptually, the operation consists of two message-passing steps: nodes first send information to their connected hyperedges, which then aggregate and redistribute this information back to their constituent nodes. This bidirectional flow enables the model to capture group-level dependencies that cannot be represented by pairwise interactions alone. Such modeling capability is especially important in supply chains involving shared products. After $L$ layers, the model generates refined firm node embeddings $\{\mathbf{z}_c^{(t)}\}_{c \in \mathcal{C}}$ that capture both product-level associations and the hierarchical structure of the supply chain. Although HGNN is used in our implementation, the architecture remains flexible and can be extended with more advanced hypergraph convolution methods such as UniGNN~\cite{huang2021unignn}, ED-HNN~\cite{wang2022equivariant}, or KHGNN~\cite{xie2025k}.

    \paragraph{Graph-Level Aggregation.}
    To obtain a compact system-level representation at time $t$, we aggregate the embeddings of all firm nodes using a permutation-invariant function:
    \begin{equation}
        \mathbf{s}_t = \mathrm{Pool}\left(\left\{\mathbf{z}_c^{(t)}\mid c\in\mathcal{C}\right\}\right)
        \label{eq:graph-aggregation}
    \end{equation}
    where \(\mathrm{Pool}(\cdot)\) can be instantiated as summation or averaging over firm nodes. The resulting $\mathbf{s}_t$ represents the structural embedding of the supply chain at time $t$, capturing both node-level features and the global hypergraph structure. Notably, product node embeddings are excluded from this aggregation because they do not represent intrinsic dynamic states. Instead, they function as intermediaries that facilitate information exchange among firms during hypergraph convolution.

    \subsection{Resilience Inference and Optimization}
    For the resilience inference, we apply a global readout layer that aggregates the structural embeddings over the temporal window:
    \begin{equation}
        \tilde{\mathbf{s}} = \mathrm{Readout}\left(\left\{\mathbf{s}_t\mid t=0,1,\dots,T-1\right\}\right)
        \label{eq:temporal-aggregation}
    \end{equation}
    where \(\mathrm{Readout}(\cdot)\) may refer to any pooling function or temporal modeling technique, such as a Transformer Encoder~\cite{vaswani2017attention}. In this study, we employ simple mean pooling to emphasize the formulation of the resilience inference task and to demonstrate the effectiveness of the hypergraph-based framework, without introducing additional complexity from advanced temporal models. Subsequently, the aggregated representation $\tilde{\mathbf{s}}$ is transformed by an MLP to generate the final resilience prediction. Finally, the entire model is optimized in an end-to-end manner using the binary cross-entropy loss.

    \section{Experiments}

    \subsection{Datasets}\label{subsec:datasets}

    We evaluate our approach on two datasets: a real-world supply chain network centered on Tesla (denoted TESLA) and a collection of synthetic networks (denoted SCR) generated using the publicly available \texttt{SupplySim} simulator~\cite{chang2025learning}. Key dataset statistics are presented in Table~\ref{tab:dataset_stats}, with detailed construction procedures provided in the Appendix.

    \begin{table}[t]
        \centering
        \setlength{\tabcolsep}{4pt} 
        \begin{tabular}{lcccccc}
            \toprule
            & \multicolumn{3}{c}{\textbf{Tesla}} & \multicolumn{3}{c}{\textbf{SCR}} \\
            \cmidrule(lr){2-4} \cmidrule(lr){5-7}
            & Train & Val & Test & Train & Val & Test \\
            \midrule
            \#Res.\      & 246   & 59  & 57   & 195   & 29  & 41   \\
            \#Non-Res.\  & 434   & 61  & 63   & 180   & 40  & 45   \\
            \#Total      & 680   & 120 & 120  & 375   & 69  & 86   \\
            \midrule
            Avg.\ \#Nodes & 96    & 98  & 111  & 55    & 57  & 53   \\
            Avg.\ \#Edges & 43    & 43  & 51   & 366   & 362 & 358  \\
            \bottomrule
        \end{tabular}
        \caption{Dataset statistics for TESLA and SCR.}
        \label{tab:dataset_stats}
    \end{table}

    \paragraph{TESLA.}
    We collect all U.S.\ import bills of lading from the ImportYeti platform that list \textit{Tesla, Inc.} as the consignee, covering the period from \textit{January 1, 2015} to \textit{June 15, 2025}. Each shipment record contains a Harmonized System (HS) commodity code, which we categorize based on the first two digits to capture high-level commodity classifications. For each HS category, we identify the direct suppliers of Tesla from the shipment data. We then extend the network by tracing downstream demand connections: initially identifying the customers of Tesla's suppliers, and subsequently identifying the customers of those customers. This process yields a three-tier, demand-centric network, with Tesla positioned as a key downstream entity. Although the dataset lacks domestic transaction records and contains incomplete maritime shipping data, it still enables a meaningful approximation of Tesla's extended supply-demand network structure.

    \paragraph{SCR.}
    To support reproducible research and evaluate model performance on complex benchmark networks, we employ the \texttt{SupplySim} simulator~\cite{chang2025learning}, which replicates the topological and transactional characteristics of real-world supply chains. We generate approximately 500 synthetic networks exhibiting diverse structural patterns. To simulate scenarios of data incompleteness and evaluate model robustness under such conditions, we create perturbed variants of test networks by applying random removals with fixed probability $p=0.15$. Specifically, we consider: (i) \textit{Node Removal} (SCR-NR), where each firm or product node is independently dropped with probability $p$, along with all incident edges; and (ii) \textit{Edge Removal} (SCR-ER), in which each edge is independently removed with probability $p$ while retaining all nodes.

    \paragraph{Resilience Label Generation.}
    We simulate inventory dynamics by combining the inventory–production feedback loop from Forrester’s system dynamics framework~\cite{forrester1997industrial} with Sterman’s exponential-smoothing demand forecast and bullwhip effect formulation~\cite{sterman2002system}. Following the ResInf~\cite{ResInf}, we sample $n=12$ random initial inventory vectors for each network and simulate 200 discrete time steps. A binary label $y \in \{0,1\}$ is assigned based on whether all trajectories converge to a common equilibrium.

    \begin{table*}[t!]
        \centering
        \begin{tabular}{lllll}
            \toprule
            \textbf{Model}                        & SCR                   & SCR-NR                & SCR-ER                & TESLA                 \\
            \midrule
            MLP                                   & 0.684(0.006)          & 0.664(0.010)          & --                    & 0.655(0.009)          \\
            \midrule
            ResInf                                & 0.499(0.176)          & 0.467(0.173)          & 0.463(0.165)          & 0.666(0.118)          \\
            GIN                                   & 0.644(0.122)          & 0.572(0.101)          & 0.656(0.036)          & 0.679(0.051)          \\
            GraphSAGE                             & 0.714(0.033)          & 0.694(0.010)          & 0.713(0.026)          & 0.806(0.036)          \\
            \midrule
            SC-RIHN
            & \textbf{0.770}(0.014)$^{**}$
            & \textbf{0.709}(0.007)$^{*}$
            & \textbf{0.811}(0.016)$^{**}$
            & \textbf{0.856}(0.017)$^{**}$ \\
            w/o Positional Embeddings & 0.727(0.005) &  0.663(0.010) & 0.744(0.009)  & 0.736(0.009) \\
            w/ product nodes & 0.737(0.005) & 0.653(0.009) & 0.771(0.016) & 0.820(0.031) \\
            \bottomrule
        \end{tabular}
        \caption{Mean F1-score (standard deviation in parentheses) over 10 random runs.
        SCR-ER is not applicable to MLP, which does not utilize edge information.
        A statistically significant improvement over the best-performing baseline is indicated with a star ($^{*}$, $p < 0.05$; $^{**}$, $p < 0.005$) according to two-sided paired t-tests.
        }
        \label{tab:main_results}
    \end{table*}

    \subsection{Experimental Setup}
    To maximize data efficiency, each supply chain network generates 12 distinct trajectory samples that share the same underlying network topology but differ in temporal dynamics. Similar to ResInf~\cite{ResInf}, only the first $T = 5$ inventory states are used to construct the historical observation window, emphasizing early-stage dynamics. We perform a disjoint split at the supply chain level, assigning all associated samples exclusively to the training, validation, or test set in proportions of approximately 70\%, 15\%, and 15\%, respectively, to prevent information leakage. Models are trained for 20 epochs, and the checkpoint with the highest validation macro-F1 on the validation set is selected for final testing. Training is conducted using the Adam~\cite{kingma2014adam} optimizer with a learning rate of 0.001 and batch size of 64. All models use a hidden dimension of 64. The number of layers $L$ is tuned over $\{2, 3, 4, 5\}$, and the pooling function is set to mean pooling. Experiments are performed on a machine running Ubuntu 22.04.1 with 4 NVIDIA RTX 4090 GPUs.

    \subsection{Baselines}

    We compare SC-RIHN against both structure-unaware method and representative GNN-based models to evaluate the benefits of hypergraph modeling:
    (1) \textbf{MLP}: Processes each firm independently without utilizing structural information;
    (2) \textbf{ResInf}~\cite{ResInf}: Combines GCN~\cite{GCN} and Transformer~\cite{vaswani2017attention}, adapted via hyperedge-to-clique expansion;
    (3) \textbf{GIN}~\cite{xu2018powerful}: A highly expressive GNN based on injective aggregation, applied to clique-expanded graphs;
    and (4) \textbf{GraphSAGE}~\cite{hamilton2017inductive}: An inductive GNN that enables representation generalization via sampled neighbor aggregation.

%

    \subsection{Main Results}

     Table~\ref{tab:main_results} shows that SC-RIHN consistently outperforms all baselines on both synthetic (SCR) and real-world (TESLA) datasets, confirming the effectiveness of explicit hypergraph modeling for supply chain resilience inference. SC‑RIHN remains robust under network perturbations and even improves slightly in edge removal scenarios, suggesting that pruning noisy or irrelevant links can enhance resilience estimation. On the dense SCR dataset, ResInf and GIN underperform relative to MLP, highlighting their limitations in capturing complex multi‑party interactions. GraphSAGE achieves better results, due to its sampling‑based local aggregation that reduces sensitivity to noisy or redundant edges. For the cleaner TESLA dataset, all GNN models surpass MLP, demonstrating their strength under well‑structured graphs. However, the elevated variance observed in ResInf and GIN indicates sensitivity to structural perturbations, which may result from information loss during the hypergraph-to-graph conversion.

    Additionally, removing positional embeddings from the feature encoder significantly degrades performance, underscoring their critical role in preserving feature-index identity. Similarly, including product nodes in graph-level pooling degrades performance, suggesting they function best as intermediaries for message propagation rather than as contributors to the final prediction.

    \subsection{Ablation Studies}\label{subsec:ablation-studies}
    \begin{figure}[h]
        \centering
        \begin{subfigure}{0.95\columnwidth}
            \centering
            \includegraphics[width=\linewidth]{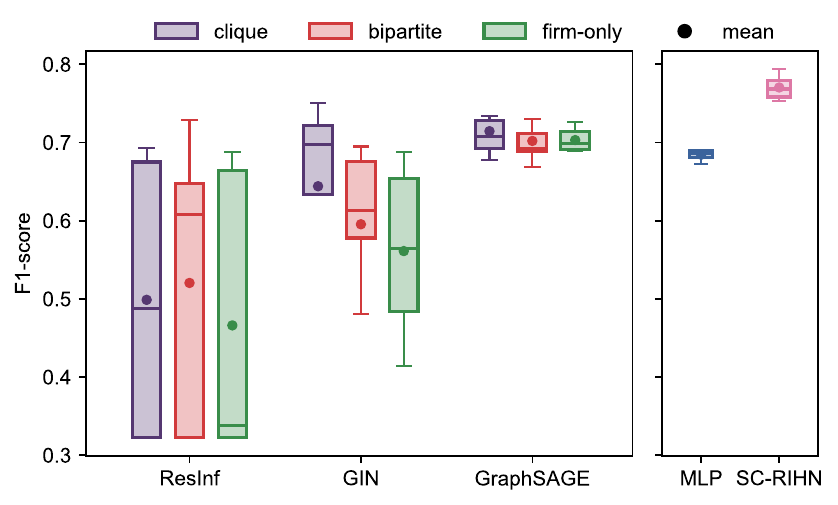}
            \caption{SCR dataset}
            \label{fig:ablation-mapping-scr}
        \end{subfigure}

        \begin{subfigure}{0.95\columnwidth}
            \centering
            \includegraphics[width=\linewidth]{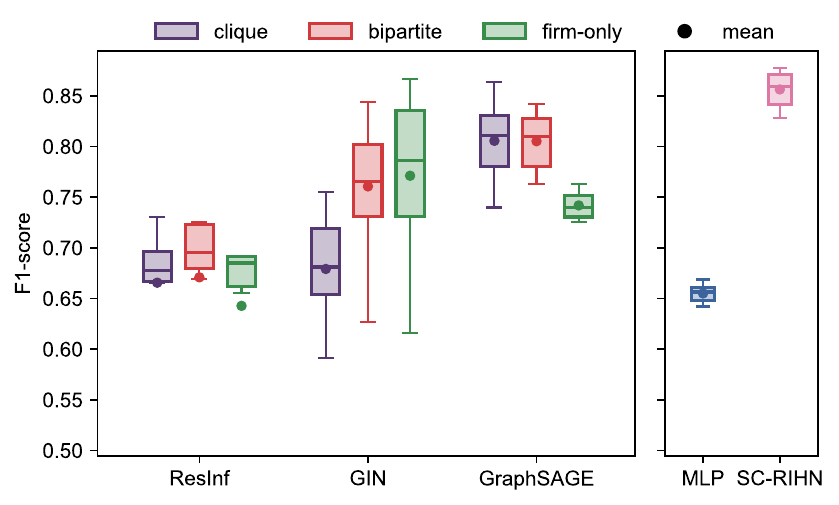}
            \caption{TESLA dataset}
            \label{fig:ablation-mapping-tesla}
        \end{subfigure}
        \caption{
        Performance of GNN baselines with different hypergraph reduction strategies on (a) SCR and (b) TESLA.
        }
        \label{fig:gnn_reduction}
    \end{figure}
    \subsubsection{Limitations of GNNs via Hypergraph Reduction.}

    To evaluate whether standard GNNs can substitute for hypergraph modeling in supply chain resilience inference, we apply three reduction heuristics: (1) Clique, which fully connects all nodes in a hyperedge; (2) Bipartite, which retains only firm–product edges; and (3) Firm-only, which removes product nodes entirely. We evaluate ResInf, GIN, and GraphSAGE on the resulting graphs. As shown in Figure~\ref{fig:gnn_reduction}, all reduced models perform worse than SC‑RIHN, confirming that flattening higher-order interactions leads to loss of structural information critical for resilience inference.

Among the three methods, Clique and Bipartite perform similarly, as both preserve firm–product links that support indirect inter-firm communication. Firm-only performs notably worse, highlighting the importance of product nodes in preserving structural pathways and enabling multi-party message exchange. A slight exception occurs with GIN on TESLA, possibly due to noise from sparse product links. GraphSAGE benefits from product-mediated patterns but still trails behind SC‑RIHN.


    \paragraph{Prediction Behavior across Resilience Classes.}

\begin{figure}[h]
        \centering
        \includegraphics[width=0.85\columnwidth]{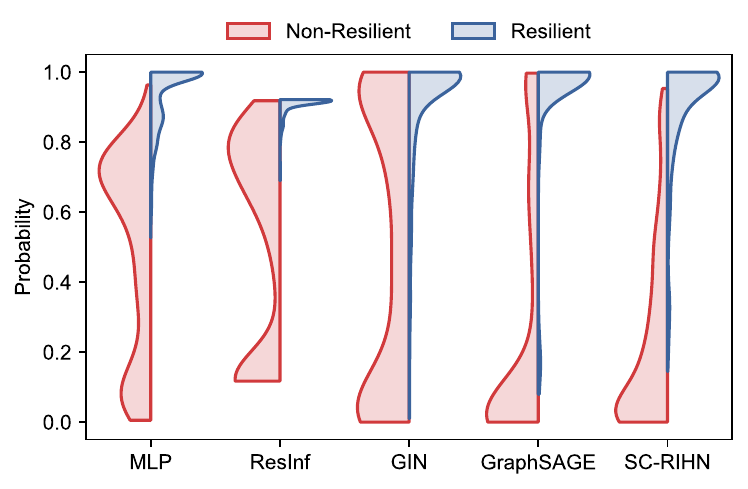}
        \caption{
            Distribution of predicted resilience probabilities on the TESLA test set.
        }
        \label{fig:violin-tesla}
    \end{figure}

    To evaluate the ability of each model to distinguish resilient from non-resilient supply chains in realistic settings, we visualize the predicted positive-class probabilities on the TESLA test set (Figure~\ref{fig:violin-tesla}). All models reliably assign high scores to resilient cases, indicating their shared capacity to detect strong resilience signals. However, distinguishing non-resilient cases remains challenging. SC-RIHN more effectively suppresses scores for these samples, while GraphSAGE shows overconfidence on a few non-resilient samples despite its overall competitiveness. In contrast, MLP, ResInf, and GIN exhibit significant overlap between the two classes, leading to higher false positives and reduced interpretability. These results underscore the difficulty of calibrated prediction in real-world supply chains and suggest that while hypergraph-based models offer improvement, there remains room for more precise uncertainty estimation.

\begin{figure}[h]
        \centering
        \includegraphics[width=0.8\columnwidth]{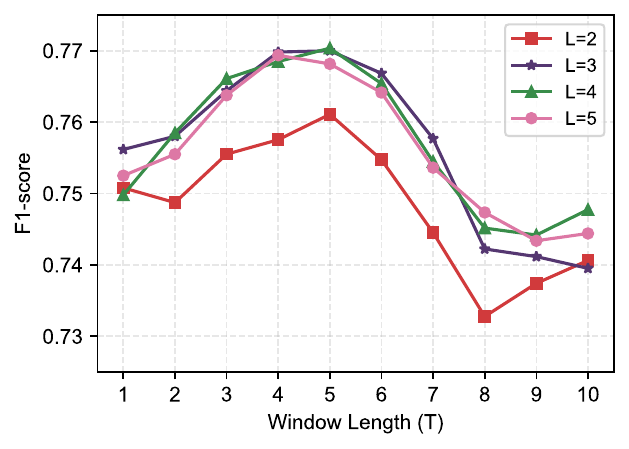}
        \caption{F1-score (mean over 10 random runs) on SCR under varying window length $T$ and SC-RIHN layer depth $L$.}
        \label{fig:f1_L_vs_W}
    \end{figure}

    \paragraph{Sensitivity to Window Length and Layer Depth.}
    We evaluate the impact of temporal context and model depth on SC-RIHN by varying the observation window length $T$ and the number of hypergraph layers $L$. As shown in Figure~\ref{fig:f1_L_vs_W}, even with a single-step observation ($T=1$), SC-RIHN achieves an F1-score around 0.75, demonstrating its ability to infer resilience from static supply chain structures by exploiting higher-order relational patterns. Performance improves with longer windows and peaks at $T=5$, beyond which it declines due to accumulated noise and the absence of explicit temporal modeling. We thus set $T=5$ as the default. Structurally, increasing the number of layers improves performance up to a point, underscoring the value of multi-hop message passing. A four-layer configuration provides the best trade-off between accuracy and complexity, while five layers occasionally degrade results due to oversmoothing.

    \section{Conclusion}
In this paper, we introduced the Supply Chain Resilience Inference (SCRI) problem, aiming to predict supply chain resilience based on hypergraph structures and historical state trajectories. We proposed the Supply Chain Resilience Inference Hypergraph Network (SC-RIHN), a novel model designed to capture complex multi-party interactions inherent to supply chains by leveraging hypergraph convolutions. Experiments conducted on synthetic and real-world datasets, including perturbation scenarios, demonstrate SC-RIHN’s superior performance over traditional graph-based models and standard machine learning baselines. Our findings highlight the critical importance of explicitly modeling higher-order dependencies, showcasing SC-RIHN’s practical value for proactive risk management and robust supply chain design. Future directions include extending our framework to dynamic temporal modeling and exploring applications across broader real-world supply chain scenarios.

\section{Acknowledgments}
This work was partially  supported by the grants of Jilin Provincial International Cooperation Key Laboratory for Super Smart City and Jilin Provincial Key Laboratory of Intelligent Policing, and the Research Institute of Trustworthy Autonomous Systems at Southern University of Science and Technology.

\bibliography{aaai2026}

\end{document}